\documentclass[letterpaper, 10 pt, conference]{ieeeconf}  

\usepackage{times}
\usepackage{epsfig}
\usepackage{epstopdf}
\usepackage{graphicx}
\usepackage{subfigure}
\usepackage{amsmath}
\usepackage{amssymb}
\usepackage{multirow}
\usepackage[lined,boxed,commentsnumbered, ruled]{algorithm2e}
\usepackage{url}

\IEEEoverridecommandlockouts                              

\title{\LARGE \bf
Place classification with a graph regularized deep neural network model
}

\author{Yiyi Liao$^{1}$, Sarath Kodagoda$^{2}$, Yue Wang$^{1}$, Lei Shi$^{2}$, Yong Liu$^{3}$ 
\thanks{$^{1}$Yiyi Liao and Yue Wang are with the Institute of Cyber-Systems and Control, Zhejiang University, Zhejiang, 310027, China.}
\thanks{$^{2}$Sarath Kodagoda and Lei Shi are with the Centre for Autonomous Systems (CAS), The University of Technology, Sydney, Australia.}
\thanks{$^{3}$Yong Liu is with the State Key Laboratory of Industrial Control Technology and Institute of Cyber-Systems and Control, Zhejiang University, Zhejiang, 310027, China (He is the corresponding author of this paper, e-mail: \small  yongliu@iipc.zju.edu.cn).}}

\begin{document}

\maketitle
\thispagestyle{empty}
\pagestyle{empty}

\begin{abstract}
Place classification is a fundamental ability that a robot should possess to carry out effective human-robot interactions. It is a nontrivial classification problem which has attracted many research. In recent years, there is a high exploitation of Artificial Intelligent algorithms in robotics applications. Inspired by the recent successes of deep learning methods, we propose an end-to-end learning approach for the place classification problem. With the deep architectures, this methodology automatically discovers features and contributes in general to higher classification accuracies. The pipeline of our approach is composed of three parts. Firstly, we construct multiple layers of laser range data to represent the environment information in different levels of granularity. Secondly, each layer of data is fed into a deep neural network model for classification, where a graph regularization is imposed to the deep architecture for keeping local consistency between adjacent samples. Finally, the predicted labels obtained from all the layers are fused based on confidence trees to maximize the overall confidence. Experimental results validate the effectiveness of our end-to-end place classification framework in which both the multi-layer structure and the graph regularization promote the classification performance. Furthermore, results show that the features automatically learned from the raw input range data can achieve competitive results to the features constructed based on statistical and geometrical information.

\end{abstract}

\section{Introduction}

Place classification is one of the important problems in human-robot interactions and mobile robotics, which aims to distinguish differences between environmental locations and assign a label (corridor, office, kitchen, etc.) to each location~\cite{yuan2011robust,pronobis2011hierarchical}. It allows robots to achieve spatial awareness through semantic understanding rather than having to rely on precise coordinates in communicating with humans. Furthermore, the semantic labels has the potential to efficiently facilitate other robotic functions such as mapping~\cite{ranganathan2011visual}, behavior-based navigation~\cite{poncela2008place}, task planning~\cite{galindo2008robot} and active object search and rescue~\cite{aydemir2011search}.

In general, place classification is carried out through environment sensing. Laser range finders, cameras and RGB-D sensors are the mostly used sensing modalities. Location and topological information can also be informative in place classification. In this work, it is attempted to exploit both the sensory data and location information. We assume all the maps in this paper contain these two parts of information and some of the maps are labeled with human knowledge. Then the place classification problem can be stated as predicting the labels of new environments given the labeled maps.


By analysing those two forms of data, sensory data and location information,  we can gain insights into the characteristics of the place classification problem. Raw sensory data encode the environment information at different locations which can provide discriminative information between different classes. However, this requires an effective feature extraction method and most of the previous works tend to extract hand-engineered features from the raw data~\cite{mozos2005supervised,sousa2007real}. Our opinion is that the hand held features, may not fully exploit the potential to achieve higher generalization ability. On the other hand, the locations encode the spatial information of the environment and indicate the local consistency of the labels, which means the positions at spatial proximity have higher probability to having the same class labels.

It is to be noted that another difficulty in place classification is the influence of different field of views (FOV) of the sensors used. For example, if a laser range finder collects 180$^\circ$ FOV data facing approximately to a corner of a corridor, it may not contain enough information for classification. If the laser range finder collects $360^\circ$ FOV data at a door of an office room, the robot might  be confused to with mixed information from two classes.

In order to address these problems, in this paper, we propose a graph regularized deep learning approach classification on multi-layer inputs.  The pipeline of our system is illustrated in Fig.~\ref{framework}, which can be split into three parts:

\begin{figure*}[tpb]
\centering
\includegraphics[height=6.2cm]{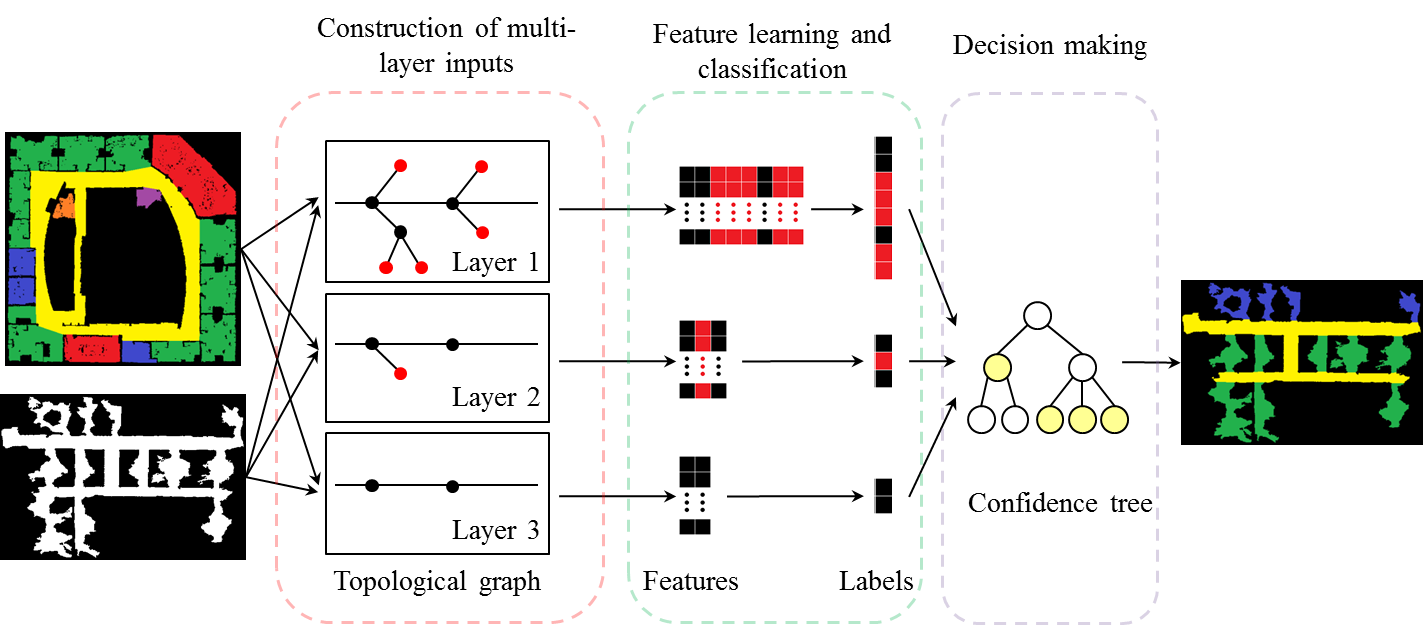}
\caption{Pipeline of the semi-supervised learning system with multi-layer inputs.}
\label{framework}
\end{figure*}

1) Construction of multi-layer inputs: The environmental information in this paper is represented through the generalized Voronoi graph (GVG)~\cite{choset1995sensor}, a \emph{topological graph} in which the nodes correspond to the sensory data and the edges denote the relationships. By fusing the information and eliminating the end-nodes, we implement a recursive algorithm to construct multi-layer inputs with hierarchical GVGs. The inputs of higher layers contain information of larger field of view, represented by increasingly succinct GVG. The features are extracted from each layer of input and classified independently.

2) The graph regularized deep architecture for feature learning and classification: We adopt the deep architecture that learns features from the raw data automatically. A graph regularizer is imposed to the deep architecture to keep the local consistency, where an \emph{adjacency graph} is constructed to depict the adjacency and similarity between the samples. Our training map and testing maps are fed into the deep architecture for feature learning at the same time, which forms a semi-supervised learning framework. The output of this step is the predicted labels of different layers.

3) The confidence tree for decision making: After receiving the classification results of multi-layer inputs, confidence trees are constructed according to the \emph{topological graph}, and a decision making process is carried out to maximize the overall confidence.

The remainder of this paper is organized as follows: Section~\ref{secRelated} reviews the related literature. In Section~\ref{secMultilayer}, we introduce the construction of our multi-layer inputs and the confidence tree for decision making. The semi-supervised classification with graph regularization is given in Section~\ref{secSemi}. Experimental results are presented in Section~\ref{secExp} to validate the effectiveness of our end-to-end classification framework. Then the paper is concluded in Section~\ref{secConclu}.

\section{Related Works} \label{secRelated}

There are various sensors that help robots to sense environments, such as cameras and laser range finders. Previous works have demonstrated the effectiveness of both camera data and laser range finder data for classifying places. For example, Shi et al.~\cite{shi2006investigating} and Viswanathan et al.~\cite{viswanathan2009automated} extracted features from the vision data, while Mozos et al.~\cite{mozos2005supervised} and Sousa et al.~\cite{sousa2007real} classified the places based on laser range data. In this paper, we focus on the place classification based on laser range data, however, our approach can be easily extended to other modality of sensors such as vision data.

Laser range finders can provide nonnegative beam sequences describing range and bearing to nearby objects. They contain structural information including clutter in the environment. Mozos et al.~\cite{mozos2005supervised} extracted features from the 360$^\circ$ laser range data and those features were fed into an Adaboost classifier to label the environment. Sousa et al. \cite{sousa2007real} reported superior results on a binary classification task using a subset of above mentioned features, and the support vector machine as the classifier. In our past work, we implemented a logistic regression based classifier, as a binary and multi-class problem contributing to higher accuracies~\cite{shi2010laser,shi2010multi}. The work was further extended to address the generalizability of the solution through a semi-supervised place classification over a generalized Voronoi graph (SPCoGVG)~\cite{shi2013towards}. In all of these methods, the features were extracted from the laser range data based on statistical and geometrical information, or so-called hand-engineered features. For instance, the average and the standard deviation of the beam length, the area and perimeter of the polygon specified by the observed range data and bearing were included in the feature set.

Recently, the unsupervised feature learning has drawn much attention as the deep learning methods was developed~\cite{HinSal06,Hinton2006,Bengio2007}. The deep learning methods achieved remarkable results in many areas, including object recognition~\cite{boureau2011ask,le2013building}, natural language processing~\cite{collobert2011natural,glorot2011domain} and speech recognition~\cite{dahl2012context}, which demonstrated that discovering and extracting features automatically can usually achieve better results on representation learning~\cite{citeulike:9426230,Rifai:2011:HOC:2034117.2034159,Vincent:2011:CSM:2000609.2000610}. Inspired by the success of unsupervised feature learning, in this article we present an end-to-end framework with the deep learning method that can learn features automatically from the laser range data.

We also exploit the local consistency of classes with the assumption that samples located in the same small region are more likely to have the same labels. Previous research has included this particular characteristic for performance promotion and many studies were carried out with consideration of the local consistency~\cite{ranganathan2011visual,pronobis2012large,mozos2005supervised,martinez2007supervised,ranganathan2010pliss}.

In this paper, we consider the local consistency during the feature learning process, where, the features learn to keep the local invariance with a graph regularization. There are some similar works on implementing the graph regularized deep learning models~\cite{Hadsell:2006:DRL:1153171.1153654,weston2012deep}. Both~\cite{Hadsell:2006:DRL:1153171.1153654} and~\cite{weston2012deep} utilized a margin-based loss function proposed by Hadsell et at.~\cite{hadsell2006dimensionality}. These works have demonstrated the effectiveness of the graph embedding in dimensionality reduction and image classification.

\section{Multi-layer Construction and Decision Making} \label{secMultilayer}

%
In this paper, we assume a laser range finder with a typical field of view of 180$^\circ$. This is a limited field of view which can give rise to many classification inaccuracies due to lack of crucial information. However, the full field of view may also lead to misclassifications at the boundaries of the two different classes of places.  Therefore, considering these problems, we propose to construct multi-layer inputs for classification followed by fusion of the results.

\subsection{Construction of Multi-layer Inputs}

\subsubsection{Data Representation on GVG}
In this paper, our multi-layer inputs is represented by the hierarchical generalized Voronoi graph (GVG)~\cite{choset1995sensor}, a topological graph which has been successfully applied to navigation, localization and mapping. The general representation of GVG is composed of meet-points (locations of three-way or more equidistance to obstacles) and edges (feasible paths between meet-points which are two-way equidistance to obstacles)~\cite{tao2011incremental}. We adopt the same resolution as in our previous work~\cite{shi2013towards} to construct the first layer GVG, and then higher layers of GVGs are constructed to describe the environment at different levels of granularity.

Let's denote hierarchical GVGs as $\langle G^{(1)},G^{(2)},\cdots,G^{(L)} \rangle$ with $G^{(l)}=\{V^{(l)},E^{(l)}\}$, where $L$ denotes the number of layer, $V^{(l)}$ denotes nodes in layer $l$ and $E^{(l)}$ denotes edges in layer $l$. For each layer, the independent sensing information is carried by the nodes in $V^{(l)}$, and the local connectivity is represented by the edges in $V^{(l)}$. More specifically, each node $v_{i}^{(l)}\in V^{(l)}$ corresponds to a sequence of range data $r_i^{(l)}$, assigned with the label $y_i^{(l)}$ for the training maps, while $e_{ij}^{(l)}\in E^{(l)}$ reveals the connection between nodes $v_i^{(l)}$ and $v_j^{(l)}$ with distance $d_{ij}^{(l)}$.

The first layer $G^{(1)}=\{V^{(1)},E^{(1)}\}$ describes the environment in most detailed level of granularity with the originally adopted laser range data. As the laser range finder is of 180$^\circ$ field of view with 1$^\circ$ angular resolution, each node $v_{i}^{(1)}\in V^{(1)}$ corresponds to a sequence of range data $r_i^{(1)}$ with 180 dimension.

\begin{algorithm}[tpb]\label{algorithm}
\caption{Generate higher layer of input from the previous layer.}
\SetKwData{Index}{Index}

\SetAlgoNoLine\LinesNumbered \KwIn{ $G^{(l)}=\{V^{(l)},E^{(l)}\}$, the range data $r_i^{(l)}$ on each node $v_i^{(l)}$}
\KwOut{ $G^{(l+1)}=\{V^{(l+1)},E^{(l+1)}\}$, the range data $r_i^{(l+1)}$ on each node $v_i^{(l+1)}$ }
\BlankLine
\For{ $v_i^{(l)}\in{V^{(l)}}$  }{
    \If { $numel(N(v_i^{(l)}))>1$ }{
    Preserve $v_i^{(l)}$, i.e. $v_i^{(l+1)}=v_i^{(l)}$;\\
    Construct $r_i^{(l+1)}$ and $\hat{r}_i^{(l+1)}$ from $r_i^{(l)}$ and all of the $r_j^{(l)}$ carried by $v_j^{(l)}\in N(v_i^{(l)})$; \\
    }
    \For{ $v_j^{(l)}\in{N(v_i^{(l)})}$  }{
    \eIf{$v_j^{(l)}\in{M(v_i^{(l)})}$}{
    Eliminate $e_{ij}^{(l)}$ and $v_{j}^{(l)}$;\\
    }
    {
    Preserve $e_{ij}^{(l)}$, i.e. $e_{ij}^{(l+1)}=e_{ij}^{(l)}$; \\
    }
    }
}
\end{algorithm}

\subsubsection{Recursive Higher Layer Construction Algorithm}

The construction of a higher layer GVG is implemented by fusing the information carried by connected nodes and then eliminating those marginal nodes. Algorithm~\ref{algorithm} demonstrates the process of building higher layer GVG from a given lower layer. We make some definitions here for better explanation of the algorithm. $N(v_i)$ is defined as the directly connected neighbour set of $v_i$, then $v_j\in N(v_i)$ means there is an edge $e_{ij}\in E$ between $v_i$ and $v_j$. In addition, $numel(N)$ is defined as the number of elements contained in $N$. Then $numel(N(v_i))=1$ means $v_i$ is an ``end-node'', i.e. the node without children. Further define $M(v_i)$ as the set of end-nodes connected to $v_i$, which is obviously $M(v_i)\subseteq N(v_i)$. As seen from Algorithm~\ref{algorithm}, the construction process fuses the information carried by $v_i$'s neighbors if $v_i$ is not an end-node (detailed fusion process is given in section \ref{secInter}), otherwise $v_i$ is eliminated from the higher layer.

\begin{figure}[tpb]
\centering
\includegraphics[height=6.5cm]{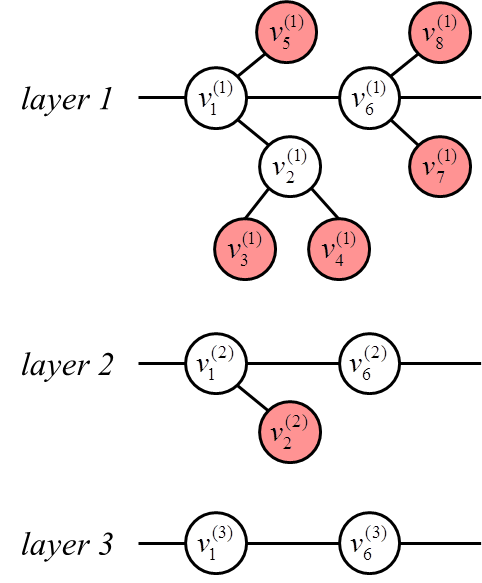}
\caption{An example of multi-layer GVG: The end-nodes are denoted as red.  The red nodes $v_1^{(1)}$, $v_2^{(1)}$ and $v_6^{(1)}$ in layer 1 are fused with their neighbours respectively, where, $v_1^{(2)}$ is composed of $(v_1^{(1)},v_2^{(1)},v_5^{(1)})$,  $v_2^{(2)}$ is composed of  $(v_1^{(1)},v_2^{(1)},v_3^{(1)},v_4^{(1)})$ and $v_6^{(2)}$ is composed of  $(v_1^{(1)}, v_6^{(1)},v_7^{(1)},v_8^{(1)})$. Then all the red nodes are eliminated from layer 1. This process will be performed recursively on layer 2 to generate layer 3. }
\label{process}
\end{figure}

The $L$ layer of data can be generated by recursively applying Algorithm~\ref{algorithm} for $L-1$ times, which means by taking the output of $l$th layer as the input of $(l+1)$th layer. This process can be illustrated in Figure~\ref{process} with $L=3$. In this example, the end-nodes are denoted as red. It is to be noted that when moving to higher layers, the number of nodes in each layer decreases with the elimination of the end-nodes. More details are given in the caption of Figure~\ref{process}.

\begin{figure*}[tpb]
\centering
\subfigure[Layer 1]{\label{gvg1}
\includegraphics[height=4.5cm]{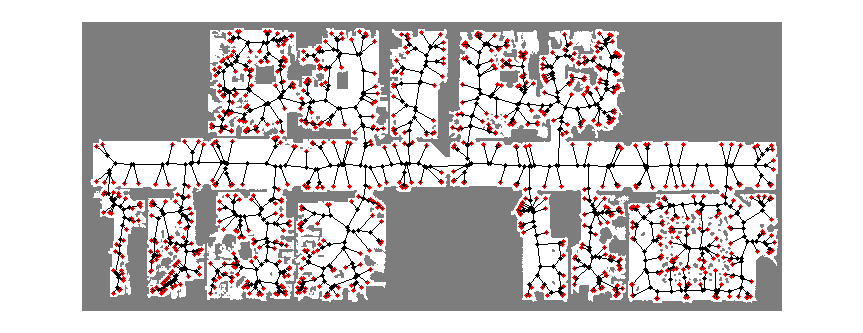}}
\subfigure[Layer 2]{\label{gvg2}
\includegraphics[height=4.5cm]{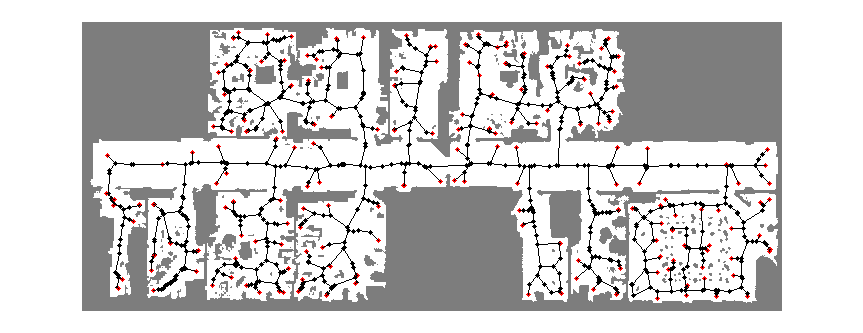}}
\subfigure[Layer 3]{\label{gvg3}
\includegraphics[height=4.5cm]{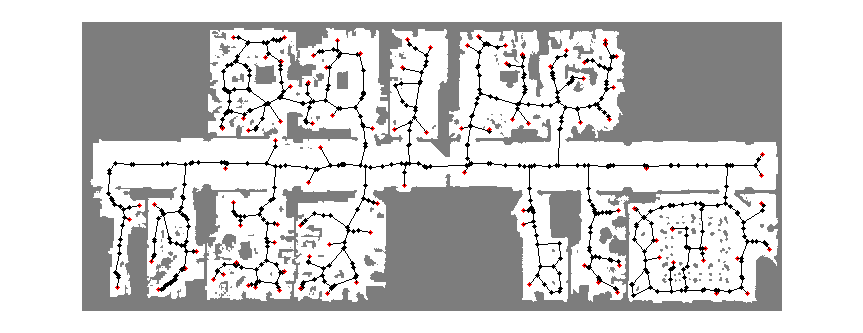}}
\caption{Multi-layer of the GVG graph $G^{(l)}=\{V^{(l)},E^{(l)}\},l=1,2,3$ on Fr79. The red nodes correspond to the end-nodes, which will be eliminated in the next layer, and the black nodes will be preserved. The edges reveals the connection between these nodes.}
\label{gvg}
\end{figure*}

An illustration of the different $G^{(l)}=\{V^{(l)},E^{(l)}\},~l=1,2,3$ layers constructed from a specific map is given in Figure~\ref{gvg}. In the first layer, the nodes are distributed more densely in the map. When approaching higher layers, the tree structure represents more and more abstract information. It is to be noted that the number of the end-nodes (denoted as red asterisks) decreases as the progression of the  layers which is a consideration for choosing the $L=3$ in our experiments.

\subsubsection{Data generation}\label{secInter}
This section describes the details about the construction of the higher-layer range data $r_i^{(l+1)}$ and $\hat{r}_i^{(l+1)}$, where the latter is generated from the former with fixed length. As stated in Algorithm \ref{algorithm}, given $v_i^{(l)}$ satisfying $numel(N(v_i^{(l)}))>1$ (i.e. $v_i^{(l)}$ is not end-node), range data received at the respective nodes are integrated to achieve a better perception.

Given each $v_i^{(l)}$ with $numel(N(v_i^{(l)}))>1$, firstly a local map is generated using occupancy grid mapping \cite{thrun2005probabilistic} based on the respective range data in $l$th layer, including $r_j^{(l)}$ carried by $v_j^{(l)}\in N(v_i^{(l)})$ and $r_i^{(l)}$. This is achieved by transforming all $r_j^{(l)}$  to $r_i^{(l)}$'s coordinate frame, which assumes the knowledge of the global robot poses at all times.  In this local map, a virtual scan $r_{i}^{(l+1)}$ is then generated by applying ray casting at position $v_i^{(l)}$ with 1$^\circ$ angular resolution, which is the same as the setting of the real laser range finder.

As the dimensions of the fused range data $r_{i}^{(l+1)}$ could be different in various nodes,  linear interpolation on the data is then performed to keep same dimension of data throughout the process. This leads to an sequence $\hat{r}_i^{(l+1)}$ with fixed dimension of 360.

Acknowledging the fact that the interpolated points may not contain high information, a completeness rate, which is the proportion of the laser measured data (dimension of $r_i^{(l)}$ to the whole 360$^\circ$ data (dimension of $\hat{r}_i^{(l)}$) is defined as:
\begin{equation}\label{complete}
q_i^{(l)} = \frac{length(r_i^{(l)})}{length(\hat{r}_i^{(l)})}
\end{equation}
where $l=2\cdots L$. This measure is used in the decision making process which is discussed in the next section, thus we denote $q_i^{(1)}=180/360=0.5$ for uniformity when $l=1$. However, we don't apply linear interpolation to the layer 1 since the initial laser range data $r_{i}^{(1)}$ always has the same dimension of 180  and is not necessary for linear interpolation.  By applying this data pre-processing approach, the laser range data in layer 2 to layer $L$ are kept in the fixed length of 360.  Note that it is always $r_{i}^{(l)}$ which is employed to construct the next layer, rather than the pre-processed $\hat{r}_{i}^{(l)}$.

\begin{figure*}[tpb]
\centering
\includegraphics[height=5.3cm]{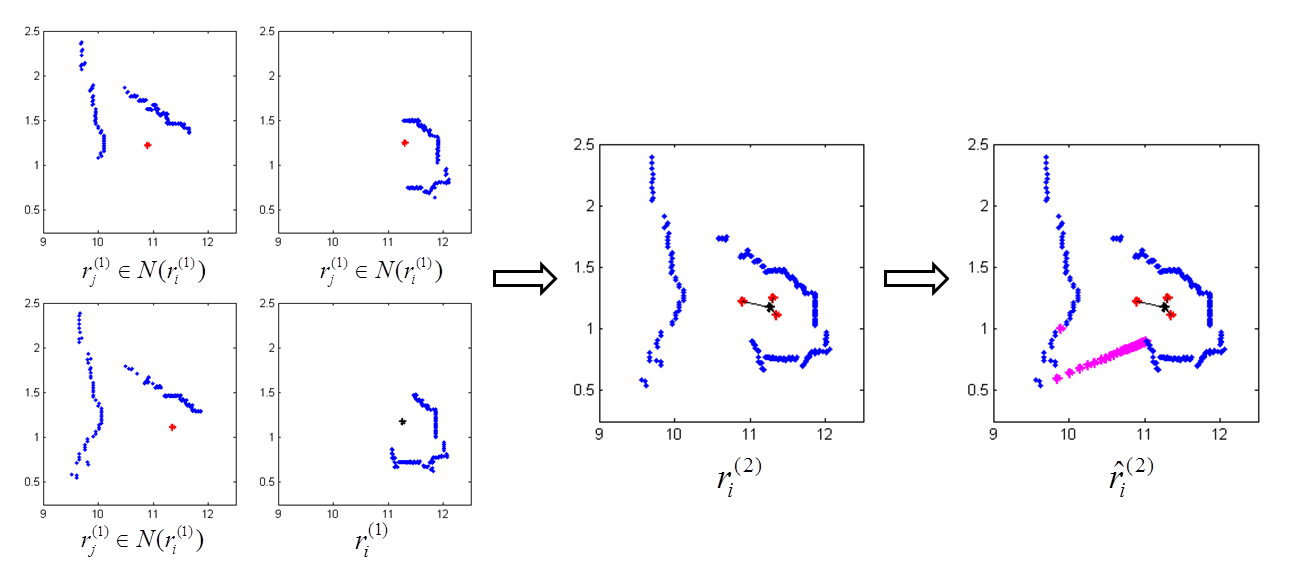}
\caption{An example of constructing $r_i^{(2)}$ and $\hat{r}_i^{(2)}$ where the axes are in meters. The left four figures illustrate $r_i^{(l)}$ and all of the $r_j^{(l)}$ carried by $v_j^{(l)}\in N(v_i^{(l)})$, where the black asterisk node denotes the position of $v_i^{(l)}$, the red asterisk nodes denote the position of $v_j^{(l)}$ and the blue nodes denote the range data collected from the real environment. Then the middle figure shows the constructed $r_i^{(2)}$ using ray casting. The interpolated sequence is given on the right, where the magenta points correspond to the interpolated ones. In this example, we have $q_i^{(2)} = 332/360 = 0.9222$.}
\label{combine}
\end{figure*}

As an example, Figure~\ref{combine} illustrates the construction of a sequence of input in layer 2 using the corresponding inputs in layer 1, followed by the result after linear interpolation.  The details are given in the caption of Figure~\ref{combine}.


\subsection{Decision on Multi-layer Results}

\subsubsection{Construction of the Confidence Tree}

With the $L$ layer of inputs, we can obtain the predicted labels from $L$ independent classifiers, which can be formed to be confidence trees with $L$ layers shown in Figure \ref{tree1}, where each node denotes the predicted label $\hat{y}_i^{(l)}$ of $v_i^{(l)}$  and its corresponding confidence $c_i^{(l)}$ . By maximizing the overall confidence of each tree structure, it is intended to obtain higher accuracy in classification.

All of these tree structures are built from the dependencies in Algorithm~\ref{algorithm} except for some minor difference --- during the construction of these tree structures, a parent node $v_i^{(l+1)}$ owns its children $v_i^{(l)}$ and $v_j^{(l)}\in M(v_i^{(l)})$, while the range data of $v_i^{(l+1)}$ is constructed from the range data carried by $v_i^{(l)}$ and $v_j^{(l)}\in N(v_i^{(l)})$. The reason is that for those nodes $v_j^{(l)}\in N(v_i^{(l)})$ and $v_j^{(l)}\notin M(v_i^{(l)})$, they are also reserved in the higher layer as  $v_j^{(l+1)}$ and have their own predicted labels, so we don't consider the influence of $v_i^{(l+1)}$ to them. It is to be noted that the number of such tree structures is equal to the number of nodes left in the layer $L$, where the $v_i^{(L)}$ are the root nodes of these trees.

In our framework, two factors are considered when computing the confidence $c_i^{(l)}$, one is the probability $p_i^{(l)}$ obtained from the classifier for labeling $\hat{y}_i^{(l)}$ and the other is the completeness ratio $q_i^{(l)}$ obtained from the input sequence $r_i^{(l)}$ which is given in (\ref{complete}). Then the confidence $c_i^{(l)}$ is constructed as:
\begin{equation}\label{confidence}
c_i^{(l)} = p_i^{(l)} \times q_i^{(l)}
\end{equation}


\subsubsection{Decision Algorithm}


\begin{algorithm}[tpb]\label{algorithm2}
\caption{Decision making on the confidence trees.}
\SetKwData{Index}{Index}

\SetAlgoNoLine\LinesNumbered \KwIn{ Confidence trees where each node $v_i^{(l)}$ denotes the predicted label $\hat{y}_{i}^{(l)}$ and the corresponding confidence $c_i^{(l)}$.}
\KwOut{ Optimized labels of leaf nodes $\hat{y}_{i*}^{(1)}$. }
\BlankLine
Initialize $c_{i*}^{(1)} = c_i^{(1)}$ and $\hat{y}_{i*}^{(1)} = \hat{y}_i^{(1)}$;\\
\For{ $l=2\cdots L$}{
\For{ $v_i^{(l)}\in{V^{(l)}}$  }{
    Average the optimized confidence of $v_i^{(l)}$'s children $v_j^{(l-1)}$ as $\frac{1}{n_i}\sum_{j}{c_{j*}^{(l-1)}}$;\\
    \eIf { $\frac{1}{n_i}\sum_{j}{c_{j*}^{(l-1)}}>c_{i}^{(l)}$ }{
    Denote $c_{i*}^{(l)}=\frac{1}{n_i}\sum_{j}{c_{j*}^{(l-1)}}$;\\
    }{
    Denote $c_{i*}^{(l)}=c_{i}^{(l)}$;\\
    All descendants of $v_i^{(l)}$ are assigned with the label $\hat{y}_{i*}^{(l)}$.
    }

}
}
\end{algorithm}

With the confidence trees denoting the predicted label $\hat{y}_i^{(l)}$  and its corresponding confidence $c_i^{(l)}$ for each given $v_i^{(l)}$, the aim of decision making is then to search the optimized confidence $c_{i*}^{(l)}$ and assign the optimized label $\hat{y}_{i*}^{(l)}$ to each node, leading to the maximum value of the overall confidence.

\begin{figure}[tpb]
\centering
\subfigure[Confidence tree]{\label{tree1}
\includegraphics[height=2.8cm]{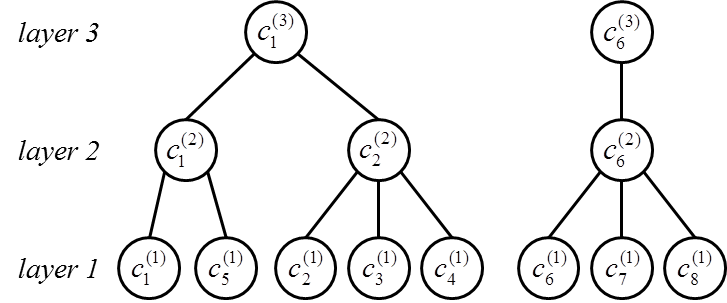}}
\subfigure[A decision example]{\label{tree2}
\includegraphics[height=2.8cm]{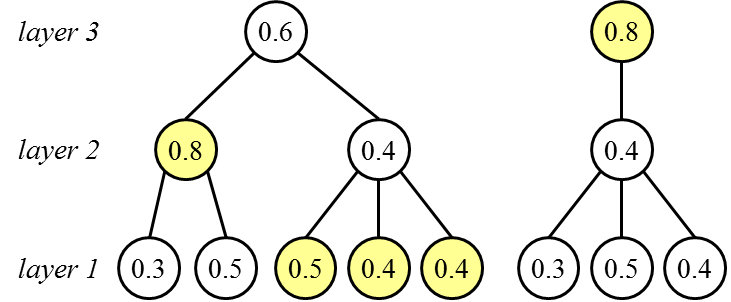}}
\caption{Confidence trees built from Figure~\ref{process} and a corresponding example. (a) The confidence tree: Each parent node $v_i^{(l+1)}$ has children $v_i^{(l)}$ and $v_j^{(l)}\in M(v_i^{(l)})$. (b) The decision example: in this example, let's assume that the confidence of each node is known. By applying the decision method given in Algorithm~\ref{algorithm2}, firstly we have the initialization $c_{i*}^{(1)} = c_i^{(1)}$ and $\hat{y}_{i*}^{(1)} = \hat{y}_i^{(1)}$. And then average confidence of the children in bottom most layer are compared with their corresponding parents in the immediate upper layer. In the left tree, $c_1^{(2)}$ is larger than the average value of $c_1^{(1)}$ and $c_5^{(1)}$, and therefore $c_{1*}^{(2)}=0.8$ and both the respective children ($v_1^{(1)}$ and $v_5^{(1)}$) are assigned the label $\hat{y}_{1}^{(2)}$. The $c_2^{(2)}$  is smaller than the average value of $c_2^{(1)},c_3^{(1)}$ and $c_4^{(1)}$, hence these leaf nodes remain their initial label and $c_{2*}^{(2)}=(0.5+0.4+0.4)/3=0.4333$. Finally $c_1^{(3)}=0.6$ is compared with $(c_{1*}^{(2)}+c_{2*}^{(2)})/2=0.6167$. Since the confidence of layer 3 is smaller than the optimized average confidence combined from layer 1 and layer 2, the final optimized confidence is $c_{1*}^{(3)}=0.6167$ and the optimized labels do not change. By applying the same decision process on the right tree in the figure, $v_6^{(1)}$, $v_7^{(1)}$ and $v_8^{(1)}$ are labeled the same as $v_6^{(3)}$.}
\label{tree}
\end{figure}

In each tree structure, we make decisions from children to parents while comparing two consecutive layers based on the decision Algorithm~\ref{algorithm2}. It is to be noted that for the comparison between layer $l$ and layer $l-1$, the confidence of the parent $v_i^{(l)}$ is always compared to the average optimized confidence of its children $v_j^{(l-1)}$ and we assume the optimized confidences in layer 1 are known as the original confidences. As for the optimized predicted labels, Algorithm~\ref{algorithm2} tells that they are only changed to follow their ancestor when this ancestor beats its children in confidence. In other words, if none ancestor of a leaf node gain advantages in confidence, then this leaf node would keep the initial label $\hat{y}_i^{(1)}$ as its optimized label $\hat{y}_{i*}^{(1)}$. Note that although we can obtain the optimized labels for all nodes from this decision algorithm, only the labels of the leaf nodes are exported as output since the classification performance is evaluated based on these leaf nodes. An example is given in Figure~\ref{tree2} for better clarity.


We can also evaluate the results obtained from those $L$ independent classifiers separately with the help of these constructed trees. To ensure the fairness, results obtained from different layer of classifiers are all compared on the accuracy of bottom layer. Obviously, the results observed from the input of layer 1 do not need to be modified while the higher layers should spread their predicted labels to the bottom layer. Given a specific layer $l~(l>1)$, all of the nodes on the bottom layer are assigned with the same label as their ancestor in layer $l$. For example, as shown in Figure~\ref{tree2}, the $v_1^{(l)},v_2^{(l)},\cdots,v_5^{(l)}$ will be labeled by the $v_1^{(3)}$'s predicted label when we evaluate the results of layer 3.

\section{Semi-supervised Learning and Classification} \label{secSemi}

We have introduced the construction of multi-layer inputs and decision making on the multi-layer results in Section \ref{secMultilayer}. In this section, we discuss the classification problem of how to train on each layer with the input data and obtain the predicted labels of the testing maps. This is implemented by a deep learning structure, with the capability to automatically learn features from the raw input data. The $L$ layer of inputs are trained through $L$ independent deep learning modes as indicated in Figure \ref{framework}, though, these models have the same structure with raw laser range data being the input and predicted labels being the output as shown in Figure \ref{model}. Thus the discussion below in this section is not confined to any specific layer and hence the superscripts are omitted. It is to be noted that our training process is semi-supervised since both the training map and the testing map are employed for model training, where only the labels of the training map are available. The semi-supervised learning process has the advantage of gaining richer information of data distribution, while keeping the spatial consistency as we will introduce in this chapter.

\subsection{Semi-supervised Learning with Graph Regularization}
In the classification problem, we denote the training  pairs as ($X_l\in\mathbb{R}^{m\times n_l}$, $Y_l\in\mathbb{R}^{1\times n_l}$) as a convention, where $m$ denotes the input dimension, $n_l$ denotes the number of training samples. Particularly, each column in $X_l$ is a sequence of laser range data $r$, i.e. $x_l^i=r_i$. The testing data can be defined in the same way as $X_u\in\mathbb{R}^{m\times n_u}$, where $n_u$ denotes the number of testing samples. Then the task of the classification problem is to obtain predicted labels of $X_u$ given $X_l$ and $Y_l$. In addition, we denote $X=[X_l~X_u]\in\mathbb{R}^{m\times n}$ as the combination of training data and testing data with $n=n_l+n_u$, since $X$ is fed into the model as a whole during our semi-supervised training process.

\begin{figure}[tpb]
\centering
\includegraphics[height=6cm]{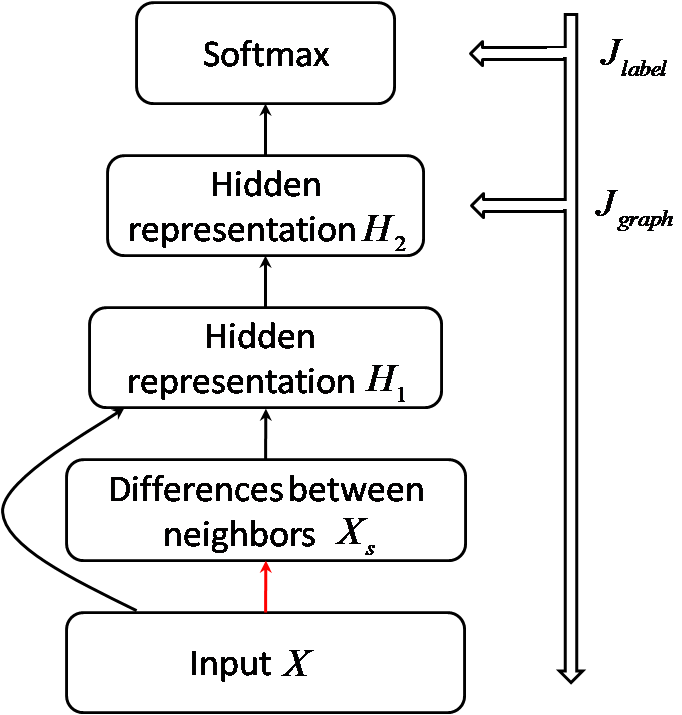}
\caption{Model training in semi-supervised learning: The second layer has fixed parameters which computes the consecutive differences of our input (denoted as red). Then both the input and the output of the second layer will be fed into the latter process. For the fine-tuning process, the $J_{label}$ is imposed to the softmax classifier and all of the parameters in the neural network (except the fixed layer) will be adjusted, while $J_{graph}$ is imposed to the last hidden layer and will only influence the feature learning process. }
\label{model}
\end{figure}

As illustrated in Figure \ref{model}, the input is firstly fed into a set of fixed parameters (denoted as red) to compute the differences between the consecutive beams in each raw scan, as the consecutive differences can also provide rich information to the place classification and is often employed for extracting geometric features in the previous works~\cite{mozos2005supervised,sousa2007real}. In the practical experiments, we sort both of the input and consecutive differences to guarantee the rotational invariance.

From this point on, both the input and output of this fixed layer are fed into the stacked auto-encoders for feature learning.  Auto-encoder is the widely used structure for building deep architectures, which is composed of an encoder and a decoder. By feeding the representation learned from the previous encoder as the input into another auto-encoder, we can obtain the stacked hidden representations as shown in Figure \ref{model}. Let's denote sigmoid function as $f(x)=1/(1+e^{-x})$, then the $i$th layer of encoder and decoder can be represented as follow:
\begin{equation}\label{dr}
\begin{split}
  H_i &= f(W_iH_{i-1}+b_i)\\
  \hat{H}_{i-1} &= f(W_i^TH_i+c_i)
\end{split}
\end{equation}
where $H_{i-1}$ and $\hat{H}_{i-1}$ denote the input and its reconstruction, $H_i$ denotes the hidden representation and $W_i,b_i,c_i$ denote the weighted parameters respectively\footnote{When $i$=1, $H_{i-1}$ is the raw input --- the combination of $X$ and its consecutive differences $X_s$.}. In this paper, the weights in each pair of encoder and decoder are tied together as shown in (\ref{dr}).

For each layer of auto-encoder, the unsupervised pre-training is applied to obtain better parameters than random initialization~\cite{HinSal06} by minimizing the reconstruction cost:
\begin{equation}\label{jcostm}
    J_{pre} = \frac{1}{m}\|H_{i-1}-\hat{H}_{i-1}\|_F^2
\end{equation}
Note that the decoder is discarded after pre-training while the encoder is preserved. The hidden representation learned by the last auto-encoder can be regarded as the feature for the input to the classifier.

In the work reported here, the softmax classifier is applied to the features learned from stacked auto-encoders for classification, which is formulated as follow:
\begin{equation}\label{softmax}
p_i = \frac{exp(w_i^Th)}{\sum_{j}{exp(w_j^Th)}}
\end{equation} 
where $p_i$ corresponds to the probability that the hidden representation vector $h$ belongs to $i$th class.

After pre-training and classification, back propagation can be used to fine-tune the whole learning process for further promotion, which means the parameters of preserved encoders and softmax are trained together. In order to keep the local consistency, we add a graph regularization term during fine-tuning to learned representation. The cost function of the fine-tuning is given as follow:
\begin{equation}\label{jcost}
\begin{split}
    J_{fine} &= J_{label}+ J_{graph}\\
      &= \frac{1}{n_l}\sum_{i=1}^{n_l}{J_{label}(x_l^i,y_l^i)} + \frac{\lambda}{n}\sum_{i=1}^{n}\sum_{j=1}^{n}{s_{ij}\|h_i-h_j\|^2}
\end{split}
\end{equation}
where the first term corresponds to the prediction error of the training data, and the second term is the graph regularization. Here $h_i$ and $h_j$ are the outputs of the last hidden layer with respect to the inputs $x_i$ and $x_j$ ($x_i$ and $x_j$ are two arbitrary columns in $X$), and  $s_{ij}$ is the similarity measurement between the samples $x_i$ and $x_j$ that connected in GVG, which is an element of the adjacency graph $S=[s_{ij}]_{n\times n}$. Figure~\ref{model} also illustrates the way our cost function work. The costs caused by the prediction error is imposed on the softmax classifier and then our graph regularization is imposed on the last hidden layer. So during the fine-tuning the $J_{label}$ will influence all of the parameters, while $J_{graph}$ will only influence the parameters for feature learning.

\subsection{Graph Regularization in Place Classification Problem}

As shown in (\ref{jcost}), the learned features $h_i$ and $h_j$ with large weight $s_{ij}$ will be pushed together with the graph regularization term. In this section, we describe the details about the construction of the adjacency graph $S$ which can be built in two steps. Firstly we define the connected relationships between samples and then calculate their weights of the connected edges.

In the place classification problem, the connected relationships in the topological graph GVG are directly employed to the adjacency graph. Then the samples with close coordinates are forced to be represented by the features with close distances. As for the weights which corresponds to the strength of the graph regularization, it is inversely associated with two distances, i.e. the distance between coordinates and the distance between the input data, which can be formulated as:
\begin{equation}\label{sij}
    s_{ij} =  \frac{\alpha}{d_{ij}} +  \frac{\beta}{\|x_i-x_j\|^2}
\end{equation}
where $\alpha$ and $\beta$ are constant weights, $d_{ij}$ denotes the Euclidean distance between the sample coordinates. The second term defines the Euclidean distance between the input data. This weighting scheme dose not only evaluate the geometrical information, but also considers the closeness between inputs. For example, given an edge that connects two nodes belonging to corridor and office respectively, although $d_{ij}$ is small, $\|x_i-x_j\|^2$ can be large. Therefore, these two nodes are not forced to be too close in the representation space however still keeps the discriminative information.

\section{Experiments} \label{secExp}

To validate the effectiveness of our end-to-end multi-layer learning system, we conduct experiments on six data sets collected from six international university indoor environments (including the Centre for Autonomous Systems at the University of Technology, Sydney, several buildings in the University of Freiburg, the German Research Centre for Artificial Intelligence in Saarbruecken, and the Intel Lab in Seattle). As we stated previously, the robot collected range data at the GVG nodes using the 2D laser range finder which has a maximum range of 30m and a horizontal field of view of 180$^\circ$.

It is to be noted that the classes defined by humans can be somewhat vague and plentiful according to the different functions of places. However, the 2D range data do not contain enough discriminative information to classify all these human-designed classes.  Therefore, after careful thinking, we consider 3 target classes as: Class 1-space designed for a small number of individuals including cubicle, office, printer room, kitchen, bathroom, stairwell and elevator; Class 2-space for group activities including meeting room and laboratory; Class 3-corridor.

Among these six data sets, two of them (Fr79 and Intellab) contain all of the 3 classes but the others contain only parts of these classes. We consider the leave-many-out training, which means one data set is utilized for training and others are used for testing. Therefore, we obtained two groups of results by training on Fr79 and Intellab respectively.

The feature learning and classification model for each layer of input is shown in Figure~\ref{model}. Given the input $X\in \mathbb{R}^{m\times n}$, the dimension configuration for our learning model is $m-m-100-24-3$, which means the consecutive differences layer has the same dimension as the input, and the dimension of our hidden layers are 100 and 24 respectively. Thus the dimension of our learned features is 24. Finally the output of our model represents a probabilistic measure of data belonging to each class. Thus the output dimension is the same as the number of our classes. In addition, since we perform the interpolation to fix the dimension of the higher layers as introduced in Section~\ref{secInter}, so we have $m=180$ for $L=1$, and $m=360$ for $L=2,3,\cdots$. In this paper, we choose $L=3$.

\subsection{Multi-layer Results without Graph Regularization}

We first conduct experiments to evaluate the performance of our multi-layer inputs. Table~\ref{intel_ng} and Table~\ref{fr79_ng} shows the leave-many-out classification results training on Intellab and Fr79 respectively. It is to be noted that the graph regularization is not considered here and therefore, $\lambda=0$ in the cost function (\ref{jcost}). In general, results of higher layers are better than that of lower layers due to the richer information contained in each node on the higher layers.

\begin{table}[tpb]
\caption{Multi-layer results trained on Intellab.}
\label{intel_ng}
\begin{center}
\begin{tabular}{cccc}
Map     &L1(\%)     &L2(\%)     &L3(\%)   \\
\hline
UTS     &85.20 	    &89.49      &91.24\\
SarrB   &86.55 	    &87.64      &91.32\\
FrUA    &86.23 	    &92.96      &91.69\\
FrUB    &90.29      &98.87      &99.84\\
Fr79    &81.99      &85.87      &87.90\\
Average     &\textbf{86.05}      &\textbf{90.97}      &\textbf{92.40}\\
\end{tabular}
\end{center}
\end{table}

\begin{table}[tpb]
\caption{Multi-layer results trained on Fr79.}
\label{fr79_ng}
\begin{center}
\begin{tabular}{cccc}
Map     &L1(\%)     &L2(\%)     &L3(\%)   \\
\hline
UTS     &81.70 	    &85.99      &89.93\\
SarrB   &84.16 	    &95.44      &90.46\\
FrUA    &90.43 	    &94.70      &96.91\\
FrUB    &88.67      &98.87      &99.51\\
Intellab    &72.55      &79.81      &82.73\\
Average     &\textbf{83.50}      &\textbf{90.96}      &\textbf{91.91}\\
\end{tabular}
\end{center}
\end{table}

\subsection{Multi-layer Results with Graph Regularization}

We also carried out experiments to validate the effectiveness of the graph regularization. The algorithm remains the same as previous settings, however, we changed  the value of  $\lambda=1$ to add the graph regularization. In this experiments, we pay more attention to the geometrical neighborhood, thus we use $\alpha=2/3$ and $\beta=1/3$ in (\ref{sij}) for the construction of the adjacency graph. The classification results are shown in Table~\ref{intel} and Table~\ref{fr79}, which are trained on Intellab and Fr79 respectively. The results have the similar trends as in Table~\ref{intel_ng} and Table~\ref{fr79_ng}, where higher layers give rise to better accuracies.  Further comparisons of Table~\ref{intel_ng} and Table~\ref{intel} show that the feature learning with graph regularization performs better than without it. It reveals that the graph regularization has the advantage of improving classification performances by keeping the local consistency.

\begin{table}[pb]
\caption{Multi-layer results trained on Intellab with graph regularization.}
\label{intel}
\begin{center}
\begin{tabular}{cccc}
Map     &L1(\%)     &L2(\%)     &L3(\%)   \\
\hline
UTS     &83.54	    &87.3	    &92.29\\
SarrB   &89.59	    &96.31      &90.89\\
FrUA    &91.48	    &91.77	    &96.68\\
FrUB    &89.97	    &99.19	    &99.84\\
Fr79    &83.96	    &86.12	    &88.65\\
Average     &\textbf{87.71}	    &\textbf{92.14}	    &\textbf{93.67}\\
\end{tabular}
\end{center}
\end{table}

\begin{table}[pb]
\caption{Multi-layer results trained on Fr79 with graph regularization.}
\label{fr79}
\begin{center}
\begin{tabular}{cccc}
Map     &L1(\%)     &L2(\%)     &L3(\%)   \\
\hline
UTS     &80.47	    &89.23	    &90.02\\
SarrB   &87.20	    &96.75      &95.23\\
FrUA    &91.06	    &96.12	    &97.47\\
FrUB    &89.48	    &98.87	    &99.51\\
Intellab    &73.00	    &79.89	    &82.51\\
Average     &\textbf{84.24}	    &\textbf{92.17}	    &\textbf{92.95}\\
\end{tabular}
\end{center}
\end{table}

\subsection{Fusion Results}
\begin{figure*}[tpb]
\centering
\subfigure[FrUA, Acc = 95.02\%]{
\includegraphics[height=4.5cm]{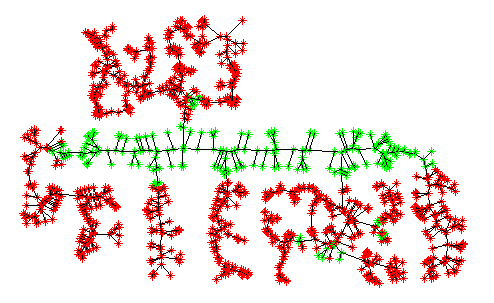}}
\subfigure[FrUB, Acc = 99.84\%]{
\includegraphics[height=4.5cm]{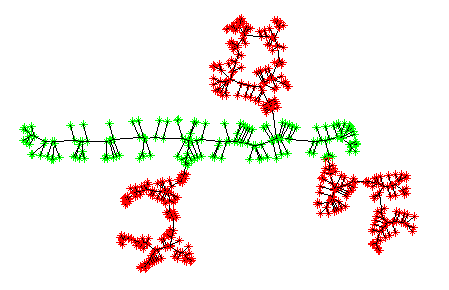}}
\subfigure[SarrB, Acc = 96.53\%]{
\includegraphics[height=4.5cm]{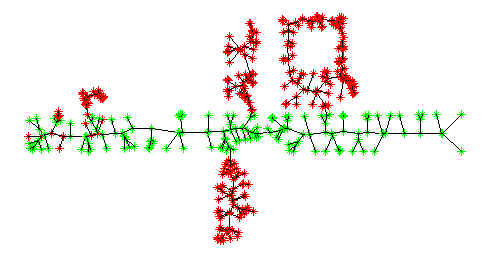}}
\subfigure[UTS, Acc = 91.24\%]{
\includegraphics[height=5cm]{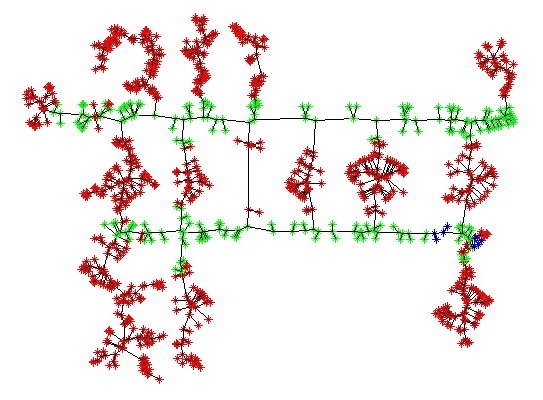}}
\subfigure[Fr79, Acc = 89.76\%]{
\includegraphics[height=4.5cm]{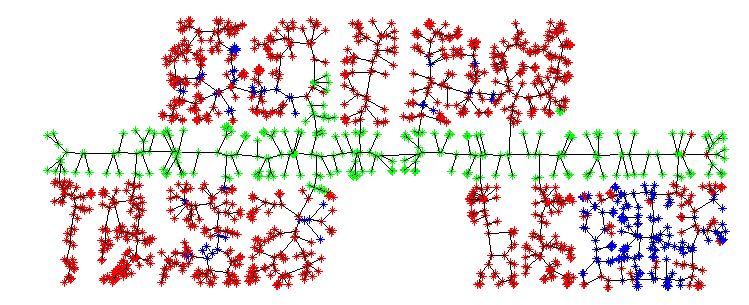}}
\subfigure{
\includegraphics[height=2.8cm]{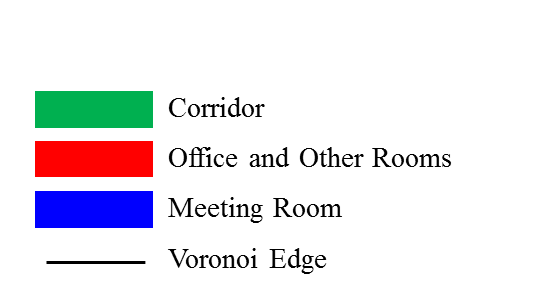}}
\caption{Test results corresponding to Table~\ref{intel_fusion}, the GVG nodes are labeled with the graph regularized fusion results trained on Intelmap.}
\label{final}
\end{figure*}

Finally, we show the accuracies of the multi-layer graph regularized method with fusion in Table~\ref{intel_fusion} and Table~\ref{fr79_fusion}. When compared with the results of each single layer as shown in Table~\ref{intel} and Table~\ref{fr79}, the fusion results achieved better accuracies. For the results trained on Intellab, the average accuracy of fusion results risen to 94.02$\%$ from  $L1$:87.71$\%$, $L2$:92.14$\%$ and $L3$:92.66$\%$, and the results trained on Fr79 also reached $93.59\%$ from $L1$:84.24$\%$, $L2$:92.17$\%$ and $L3$:92.95$\%$. The fused test results trained on Intellab are diagrammatically illustrated in Figure \ref{final}. It is to be noted that confusions between Class 1 (office room and other rooms) and Class 2 (meeting room) account for the major misclassifications especially in the test map of Fr79. The cause might be that Class 1 is featured with narrow environment including massive clutters while the Class 2 is featured with relatively larger spaces, therefore the corners of meeting room are mostly classified as office room and other rooms and some center positions of office room are assigned as office room.

We also make comparisons with the results we achieved in our previous work SPCoGVG~\cite{shi2013towards}. SPCoGVG is also a semi-supervised approach, which is composed of support vector machine (SVM) and conditional random field (CRF) to ensure the generalization ability. We use the 24-dimensional hand-engineered features in SPCoGVG, which are extracted from the raw range data with geometrical knowledge. Notice that our learned features have the same dimension as the hand-engineered features in our experiments. Seen from Table~\ref{intel_fusion} and Table~\ref{fr79_fusion}, we achieve slightly better average results than SPCoGVG.
\begin{table}[tpb]
\caption{Graph regularized fusion results trained on Intellab and results using SPCoGVG.}
\label{intel_fusion}
\begin{center}
\begin{tabular}{ccc}
Map     &Multi-layer fusion(\%)     &SPCoGVG(\%)        \\
\hline
UTS     &91.24	    &90.72	    \\
SarrB   &96.53	    &88.72      \\
FrUA    &95.02	    &96.52	    \\
FrUB    &99.84	    &98.71	    \\
Fr79    &89.76	    &92.04	    \\
Average     &\textbf{94.48}	    &\textbf{93.39}	    \\
\end{tabular}
\end{center}
\end{table}

\begin{table}[tpb]
\caption{Graph regularized fusion results trained on Fr79 and results using SPCoGVG.}
\label{fr79_fusion}
\begin{center}
\begin{tabular}{ccc}
Map     &Multi-layer fusion(\%)     &SPCoGVG(\%)        \\
\hline
UTS     &90.54	    &89.84	    \\
SarrB   &98.27	    &93.71      \\
FrUA    &97.23	    &97.71	    \\
FrUB    &99.51	    &99.19	    \\
Intellab    &82.40	    &86.89	    \\
Average     &\textbf{93.59}	    &\textbf{93.47}	    \\
\end{tabular}
\end{center}
\end{table}

\section{Conclusions}\label{secConclu}
In this paper, we presented an end-to-end place classification framework. We implemented a multi-layer learning framework, including the construction of multi-layer inputs and decision making on the multi-layer results. Each layer of inputs were fed into a semi-supervised model for feature learning and classification, which guaranteed the local consistency with a graph regularization.

Experimental results showed that the higher layer input data led to higher classification accuracy, which validated the effectiveness of the multi-layer structure. By performing the semi-supervised learning with or without graph regularization, we also showed that graph regularization help promoting the classification performance by keeping the local consistency. Furthermore, the fusion results based on the confidence tree achieved comparable results to the state-of-art method. In a nutshell, we achieved the generalization ability and preserved the local consistency in our end-to-end place classification framework. Future work is to apply our framework on other type of sensor data, such as RGB-D data, which have more representative and discriminative ability.


\bibliographystyle{ieee}
\bibliography{rootbib}

\end{document}